%% file: 2-main_regular.tex
\documentclass[9pt]{article}
\usepackage[preprint]{spconf}
\usepackage{amsmath,graphicx}
\usepackage{amsfonts}
\usepackage{bm}
\usepackage{hyperref}
\usepackage{url}
\usepackage{graphicx}
\usepackage{wrapfig}
\usepackage{booktabs}
\usepackage{multirow}
\usepackage{booktabs,arydshln}
\usepackage{floatrow}
\usepackage{enumitem}
\usepackage{floatrow}
\usepackage{sidecap}
\usepackage{wrapfig,lipsum,booktabs}
\usepackage{subcaption}
\usepackage[tableposition=top]{caption}

\usepackage[subtle]{savetrees}
\usepackage{setspace}
\input{macros}

\title{Audio-Visual Neural Syntax Acquisition}

\name{%
\begin{minipage}{\linewidth}
\centering
\emph{Cheng-I Jeff Lai\textsuperscript{1*}, Freda Shi\textsuperscript{2*}, Puyuan Peng\textsuperscript{3*},} \\
\emph{Yoon Kim\textsuperscript{1}, Kevin Gimpel\textsuperscript{2}, Shiyu Chang\textsuperscript{4}, Yung-Sung Chuang\textsuperscript{1}, Saurabhchand Bhati\textsuperscript{1},} \\
\emph{David Cox\textsuperscript{5}, David Harwath\textsuperscript{3}, Yang Zhang\textsuperscript{5}, Karen Livescu\textsuperscript{2}, James Glass\textsuperscript{1}\thanks{\hspace{-1mm}* First three authors contributed equally. Correspond to \texttt{clai24@mit.edu} and \texttt{freda@ttic.edu}}}
\end{minipage}
}

\address{
  $^1$MIT\hspace{2mm} 
  $^2$TTIC \hspace{2mm} 
  $^3$UT Austin \hspace{2mm} 
  $^4$UC Santa Barbara \hspace{2mm} 
  $^5$MIT-IBM Watson AI Lab \\
  \url{https://github.com/jefflai108/AV-NSL}
} 
\begin{document}
\maketitle
\begin{abstract}
We study phrase structure induction from visually-grounded speech.
The core idea is to first segment the speech waveform into sequences of word segments, and subsequently induce phrase structure using the inferred segment-level continuous representations.
We present the Audio-Visual Neural Syntax Learner (AV-NSL) that learns phrase structure by listening to audio and looking at images, without ever being exposed to text.
By training on paired images and spoken captions, AV-NSL exhibits the capability to infer meaningful phrase structures that are comparable to those derived by naturally-supervised text parsers, for both English and German.
Our findings extend prior work in unsupervised language acquisition from speech and grounded grammar induction, and present one approach to bridge the gap between the two topics.
\end{abstract}
\input{src/10-intro}
\input{src/20-related}

\input{src/30-method}
\input{src/40-experiments}
\input{src/50-analysis}
\input{src/60-conclusion}
\newpage
\bibliographystyle{IEEEbib}
\bibliography{strings,refs}

\end{document}

%% file: macros.tex
\newcommand{\combine}{\textit{combine}}
\newcommand{\score}{\textit{score}}

\newcommand{\interalia}[1]{\cite[\textit{inter alia}]{#1}}

\makeatletter
\def\adl@drawiv#1#2#3{%
        \hskip.5\tabcolsep
        \xleaders#3{#2.5\@tempdimb #1{1}#2.5\@tempdimb}%
                #2\z@ plus1fil minus1fil\relax
        \hskip.5\tabcolsep}
\newcommand{\cdashlinelr}[1]{%
  \noalign{\vskip\aboverulesep
           \global\let\@dashdrawstore\adl@draw
           \global\let\adl@draw\adl@drawiv}
  \cdashline{#1}
  \noalign{\global\let\adl@draw\@dashdrawstore
           \vskip\belowrulesep}}
\makeatother

\def\eqref#1{equation~\ref{#1}}

\def\1{\bm{1}}

\DeclareMathAlphabet{\mathsfit}{\encodingdefault}{\sfdefault}{m}{sl}
\SetMathAlphabet{\mathsfit}{bold}{\encodingdefault}{\sfdefault}{bx}{n}



%% file: src/10-intro.tex
\vspace{-2mm}
\section{Introduction}
\vspace{-2.5mm}
Multiple levels of early language acquisition happen without supervisory feedback \cite{dupoux2018cognitive}; it is therefore interesting to consider whether automatic learning of language, from identifying lower-level phones or words to inducing high-level linguistic structure like grammar, can also be done in \textit{natural settings}. 
In these settings, we have access to parallel data from different modalities, while the amount of data is limited. 
To this end, two concurrent lines of effort have been pursued: 
\begin{itemize}[leftmargin=*,topsep=2pt]
    \setlength{\itemsep}{0pt}
    \item Zero-resource speech processing, exemplified by the unsupervised discovery of sub-phones, phones, and words \cite{jansen2013summary}, involves constructing speech models without relying on textual intermediates, and models how children naturally learn to speak prior to acquiring reading or writing skills.
    \item Grammar induction is a process that learns latent syntactic structures, such as constituency \cite{klein2002generative} and dependency trees \cite{klein2004corpus}, without relying on annotated structures as supervision. 
\end{itemize}

In recent years, multi-modal learning has emerged as a promising and effective objective in various domains: in speech processing, \cite{harwath2018learning} proposes leveraging parallel image-speech data to acquire associated words \cite{harwath2017learning} and phones \cite{harwath2019learning}; in syntax induction, \cite{shi2019visually} proposes to induce constituency parses from captioned images. 
These successes, coupled with insights from developmental psychology \cite{dupoux2018cognitive}, motivate us to develop a computational model that utilizes the visual modality to acquire both low-level words and high-level phrase structures directly from speech waveforms, without relying on intermediate text or any form of direct supervision.

\begin{figure}[t]
\centering
\includegraphics[width=1.0\linewidth]{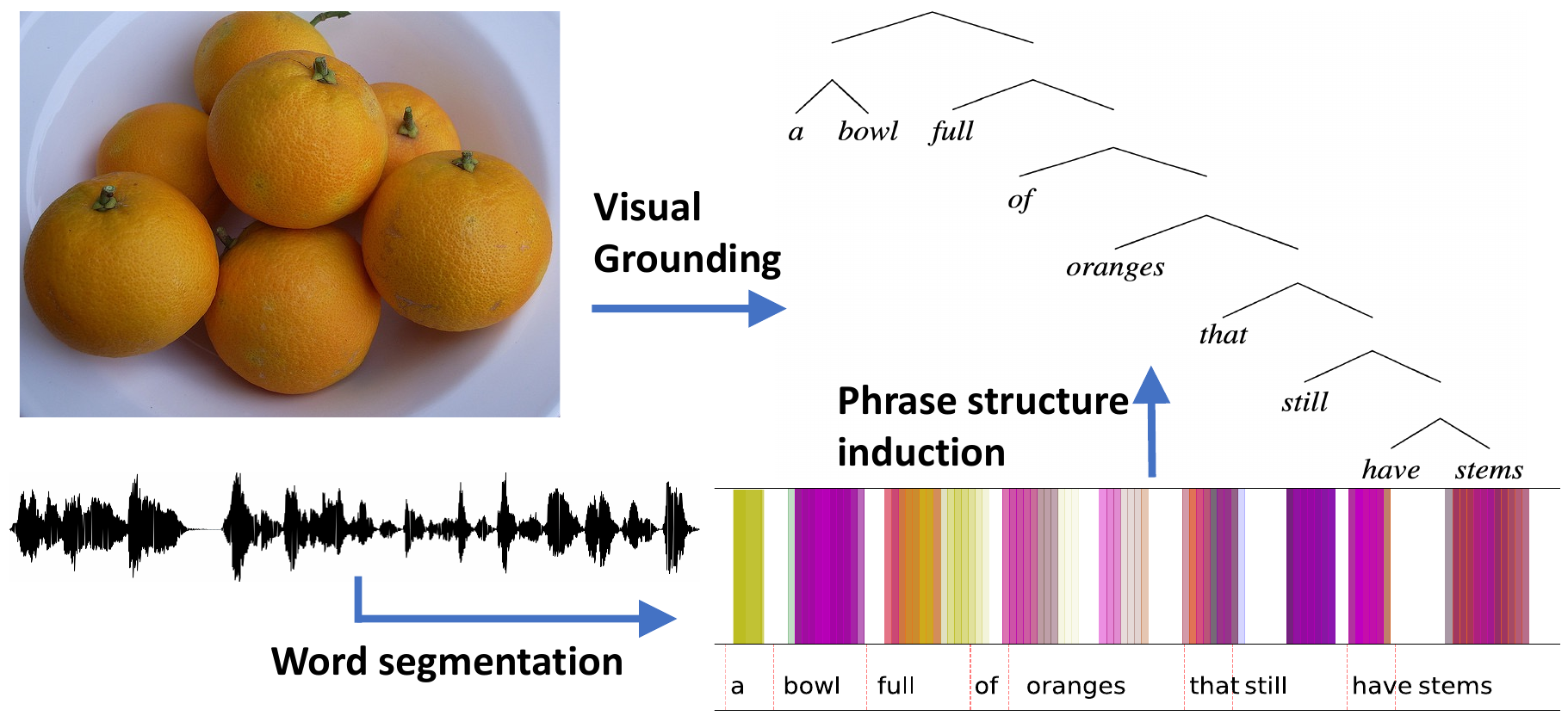}
\caption{We study the process of inducing constituency parse trees on unsupervised inferred word segments from raw speech waveforms. 
No intermediate text tokens or automatic speech recognition (ASR) is needed. 
For illustration, here we show the gold parse tree from the given text caption.}
\label{fig:teaser}
\end{figure}

In this paper, we present the Audio-Visual Neural Syntax Learner (AV-NSL; Fig.~\ref{fig:teaser}), which induces the syntactic structure of visually grounded speech utterances.  
The speech utterances are represented by sequences of \textit{continuous} speech segment representations, which are derived from a pretrained model that simultaneously discovers word-like units and learns segment representations \cite{peng2022word}.
AV-NSL (1) learns to map the representations of speech segments and images into a shared embedding space, resulting in higher similarity scores for segments and images that convey similar meanings, (2) estimates the visual \textit{concreteness} of speech segments using the learned embedding space, and (3) outputs speech segments with higher concreteness as the constituents. 

To assess the effectiveness of AV-NSL, we compare it with both the ground truth and the grounded text parser VG-NSL \cite{shi2019visually}, as well as several alternative modeling choices such as compound-PCFGs \cite{kim2019compound} over acoustic units.
An ablation study supports the reasonability of our approach. 
As a by-product, we improve over the previous state of the art in unsupervised word segmentation.

%% file: src/20-related.tex
\section{Related Work}

\noindent\textbf{Grounded grammar induction.}
Since the proposal of the visually grounded grammar induction task \cite{shi2019visually}, there has been subsequent research on the topic \interalia{zhao2020visually,zhang2021video,wan2021unsupervised}.
To the best of our knowledge, existing work on grammar induction from distant supervision has been based almost exclusively on text input. 
The most relevant work to ours is \cite{zhang2021video}, where speech features are treated as auxiliary input for video-text grammar induction; that is, \cite{zhang2021video} still requires text data and an off-the-shelf automatic speech recognition model. 
In contrast to existing approaches, AV-NSL employs raw speech data and bypasses text to induce constituency parse trees, utilizing distant supervision from parallel audio-visual data. 
\vspace{2pt}

\vspace{1mm}\noindent\textbf{Spoken word discovery.}
Following the pioneering work in spoken term discovery \cite{park2007unsupervised}, a line of work has been done to discover repetitive patterns or keywords from unannotated speech \interalia{jansen2011efficient, mcinnes2011unsupervised, zhang2013unsupervised}.
Other related work has considered tasks such as unsupervised word segmentation and spoken term discovery \interalia{lee2015unsupervised,kamper2017segmental,chorowski2021aligned, bhati2021segmental}, and the ZeroSpeech challenges \cite{dunbar2022self} have been a major driving force in the field. 
In a new line of work, \interalia{harwath2017learning} show that word-like and phone-like units can be acquired from speech by analyzing audio-visual retrieval models. 
\cite{peng2022word} shows that word discovery naturally emerges from a visually grounded, self-supervised speech model, by analyzing the model's self-attention heads.  
In contrast, AV-NSL attempts to induce phrase structure, in the form of constituency parsing on top of unsupervised word segments.
\vspace{2pt}

\vspace{1mm}\noindent\textbf{Speech parsing and its applications.}
Early work on speech parsing can be traced back to SParseval \cite{roark2006sparseval}, a toolkit that evaluates text parsers given potentially errorful ASR output. 
In the past, syntax has also been studied in the context of speech prosody \cite{wagner2010experimental, kohn2018empirical}, and \cite{tran2017parsing, tran2020role, tran2021assessing} incorporate acoustic-prosodic features for text parsing with auxiliary speech input.
\cite{lou2019neural} trains a text parser \cite{kitaev-klein-2018-constituency} to detect speech disfluencies, and \cite{pupier2022end} trains a text dependency parser from speech jointly with an ASR model. 
There is concurrent work \cite{tseng2023cascading} that extends DIORA \cite{drozdov2019unsupervised} to unsupervised speech parsing.
On the application side, syntactic parses of text have been applied to prosody modeling in end-to-end text-to-speech \cite{guo2019exploiting, tyagi2019dynamic, kaiki2021using}.
While this work builds upon pre-existing text parsing algorithms, we focus on phrase structure induction in the absence of text.

%% file: src/30-method.tex
\vspace{-2mm}
\section{Method}
\label{sec: method}
\vspace{-2mm}

\begin{figure*}[t]
\centering
\includegraphics[width=0.95\linewidth]{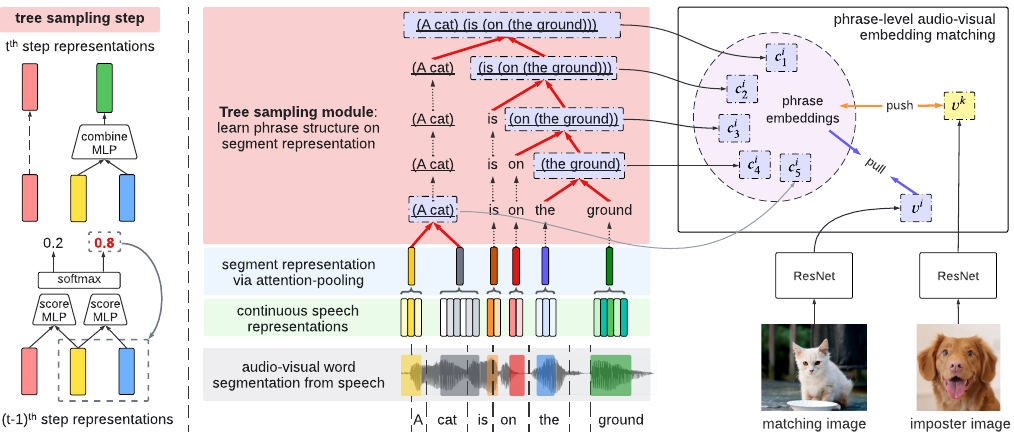}

\caption{Illustration of AV-NSL, which extends VG-NSL \cite{shi2019visually} to audio-visual inputs. 
Taking a pair of speech utterance and its corresponding image as the input, AV-NSL encodes spans of speech utterances and images into a joint embedding space. We train AV-NSL by encouraging it to output more visually concrete spans as constituents. 
Note that no text is used throughout. 
}
\label{fig:overview}
\end{figure*}

Given a set of paired spoken captions and images, the Audio-Visual Neural Syntax Learner (AV-NSL) infers phrase structures from speech utterances without relying on text.
The basis of AV-NSL is the Visually-Grounded Neural Syntax Learner (VG-NSL) \cite[\S\ref{subsec: vgnsl}]{shi2019visually}, which learns to induce constituency parse trees by guiding a sequential sampling process with text-image matching. 
We break down the problem into two steps: (1) obtaining sequences of word segments, and (2) extracting segment-level self-supervised representations.  
With these simple extensions to VG-NSL, AV-NSL induces phrase structure without reading text, but rather by listening to speech and looking at images.

\subsection{Background: VG-NSL}
\label{subsec: vgnsl}
VG-NSL~\cite{shi2019visually} consists of a bottom-up text parser and a text-image embedding matching module. 
The parser consists of an embedding similarity scoring function $\score$ and an embedding combination function $\combine$. 
Given a text caption, denoted by a sequence of word embeddings $W = \{w_i^0\}_{i=1}^N$ of length $N$, the parser synthesizes a constituency parse tree by recursively scoring and combining adjacent embeddings at each step. 
At step $t$, VG-NSL (1) evaluates all consecutive pairs of embeddings $\langle w_i^t ,w_{i+1}^t\rangle$ and assigns a scalar score to each with $\score_\Theta$, (2) selects a pair $\langle w_{i'}^t ,w_{i'+1}^t\rangle$ 
based on the corresponding scores,\footnote{In the training stage, the pair is sampled from a distribution where the probability of a pair is proportional to $\exp(\score)$; in the inference stage, the $\arg\max$ is selected.} and (3) combines the selected pair of embeddings via $\combine$ to form a new phrase embedding for the next step, copying the remaining ones to the next step.
In VG-NSL, $\score$ is parameterized by a 2-layer ReLU-activated MLP, and $\combine$ is defined by the $L_2$-normalized vector addition of the input embeddings. 
The resulting tree is inherently binary and there are $N-1$ combining steps in total, as the parser must combine two nodes in each step. 

VG-NSL trains the word embeddings $W$ and a text-image embedding matching module (parameterized with $\Phi$) jointly by minimizing the phrase-level hinge-based triplet loss: 
\begin{align*}
    \mathcal{L}_{\Phi,W} = \sum_{\mathbf{c}_W, \mathbf{i}_\Phi, \mathbf{c}'_W} & \left[\cos(\mathbf{i}_\Phi,\mathbf{c}'_W) - \cos(\mathbf{i}_\Phi, \mathbf{c}_W) + \delta\right]_+\\
    + \sum_{\mathbf{c}_W, \mathbf{i}_\Phi, \mathbf{i}'_\Phi} & \left[\cos(\mathbf{i}'_\Phi,\mathbf{c}_W) - \cos(\mathbf{i}_\Phi, \mathbf{c}_W) + \delta\right]_+,
\end{align*}
where $\mathbf{c}$, $\mathbf{i}$ are the corresponding vector representations to a pair of parallel text constituent and image; $\mathbf{c}'$ is the representation of an imposter constituent that is not paired with $\mathbf{i}$; $\mathbf{i}'$ is an imposter image representation that is not in parallel with $c$; $\delta$ is a constant margin; $[\cdot]_+ := \max(\cdot, 0)$.
By minimizing the above loss function, the embedding space brings semantically similar image and text span representations closer to each other, while pushing apart those that are semantically different.
Additionally, the loss function can be adapted to estimate the visual \textit{concreteness} of a text span: 
intuitively, the smaller the loss related to a candidate constituent $c$, the larger the concreteness of $c$, and vice versa. 
Taking the additive inverse of values inside both $[\cdot]_+$ operators, the concreteness of a constituent $c$ is defined as 
\begin{align*}
\textit{concrete}\left(\mathbf{c};\mathbf{i}\right) & =
\sum_{\mathbf{c}'}\left[\cos\left(\mathbf{i}, \mathbf{c}\right) - \cos\left(\mathbf{i}, \mathbf{c}'\right) - \delta \right]_+ \\
& + 
\sum_{\mathbf{i}'}\left[\cos\left(\mathbf{i'}, \mathbf{c}\right) - \cos\left(\mathbf{i'}, \mathbf{c}\right) - \delta \right]_+, 
\end{align*}
Finally, the estimated concreteness scores are passed back to the parser as rewards to the constituents. 
VG-NSL jointly optimizes the visual-semantic embedding loss, and trains the parser with REINFORCE \cite{williams1992simple}. 

\subsection{Audio-Visual Neural Syntax Learner}
\label{subsec: avnsl}
AV-NSL extends VG-NSL by: (1) incorporating audio-visual word segmentation to obtain sequences of word segments from unannotated speech, 
(2) jointly optimizing segment-level embeddings and phrase structure induction, and 
(3) employing deeper parameterization for the $\score$ and $\combine$ functions in the parser to handle the noisier speech representations. 
In AV-NSL, $\score$ and $\combine$ are parameterized by  GELU-activated \cite{hendrycks2016gaussian} multi-layer perceptrons (MLPs).
Below we describe (1) and (2) in detail. 

\begin{figure}[t]
\centering\strut
\includegraphics[width=1\linewidth]{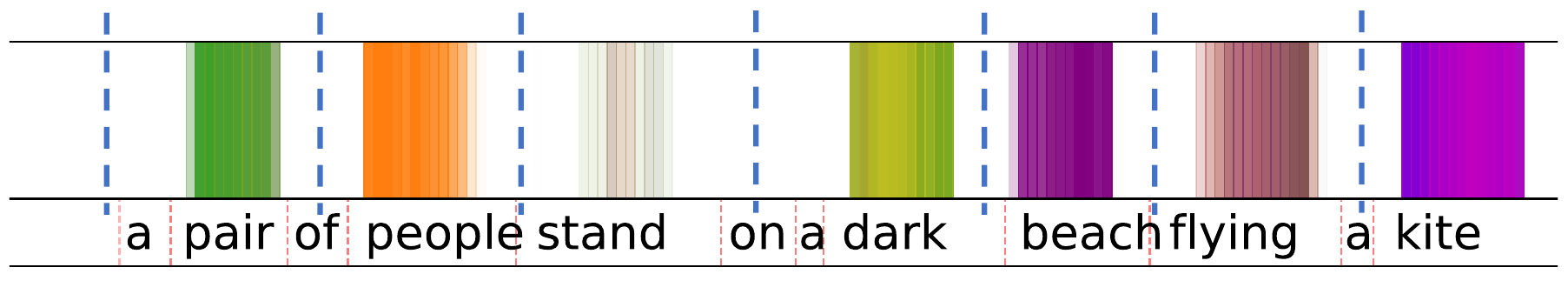}
\includegraphics[width=1\linewidth]{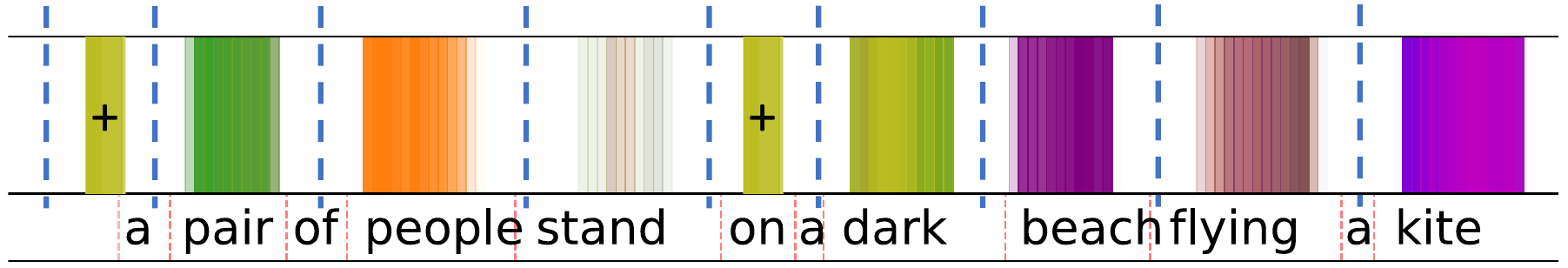}
\caption{
Example of VG-HuBERT word segmentation (top). 
Different colors denote different attention heads, and color transparency represents the magnitude of the attention weights.
Adjacent attention boundaries (vertical dashed lines) are used as the word boundaries. 
\textit{Segment insertion} (bottom): short segments (marked with ``+'') are placed in long enough gaps between existing segments to recover function words. 
Best viewed in color.
}
\label{fig:vghubert_insertion_demo}
\end{figure}

\vspace{2mm}
\noindent \textbf{Audio-visual word segmentation:}  For word segmentation,
AV-NSL leverages VG-HuBERT \cite{peng2022word} (Fig.~\ref{fig:overview}; bottom), a model trained to associate spoken captions with natural images via retrieval.
After training, spoken word segmentation emerges via magnitude thresholding of the self-attention heads of the audio encoder: at layer $l$, we (1) sort the attention weights from the \texttt{[CLS]} token to other tokens in descending order, and (2) apply a threshold $p$ to retain the top $p\%$ of the overall attention magnitude (Fig.~\ref{fig:vghubert_insertion_demo}, top).

Empirically, however, the VG-HuBERT word segmenter tends to ignore function words such as \textit{a} and \textit{of}.
Therefore, we devise a simple heuristic to pick up function word segments by inserting a short word segment
wherever there is a gap of more than $s$ seconds that VG-HuBERT fails to place a segment (Fig.~\ref{fig:vghubert_insertion_demo}). 
We additionally apply unsupervised voice activity detection \cite{tan2020rvad} to restrict segment insertion to only voiced regions.
The length of the insertion gap $s$, the VG-HuBERT segmentation layer $l$, attention magnitude threshold $p\%$, and model training snapshots across random seeds and training steps, are all chosen in an unsupervised fashion using minimal Bayes' risk decoding (\S\ref{subsec: unsup_decoding}).

\vspace{5mm}\noindent \textbf{Speech segment representations:}
We use the word segments output by VG-HuBERT to calculate the representations.
Let $R =\{r_j\}_{j=1}^T$ denote the frame-level representation sequence, where $T$ is the speech sequence length.
Audio-visual word segmentation returns an alignment $A(i) = r_{p:q}$ that maps the $i^\textit{th}$ word segment to the $p^\textit{th}$ to $q^\textit{th}$ acoustic frames.
The segment-level continuous representation for the $i^\textit{th}$ word is 
$w_i^0 = \sum_{t\in A(i)}a_{i,t}r_{i,t}$, 
where $a_{i,t}$ is the attention weights over the segments specified by $A(i)$.
In AV-NSL, $R$ is the layer representation from a pretrained speech model (e.g., VG-HuBERT), and $a_{i, t}$ is the \texttt{[CLS]} token attention weights over frames within each segment. 

\subsection{Self-Training with s-Benepar}
\label{subsec: self_train}

\cite{shi-etal-2020-role} has shown that self-training can usually improve parsing performance: the approach involves training an additional parser to fit the output generated by a pre-existing learned parser.
Concretely, \cite{shi-etal-2020-role} uses Benepar \cite{kitaev-klein-2018-constituency}, a supervised neural constituency parser, as the base model for self-training, where it (1) takes a sentence as the input, (2) maps it to word representations, and (3) predicts a score for all spans of being in the constituency parse tree. 
For inference, the model evaluates all possible tree structures and outputs the highest-scoring one.

Following \cite{shi-etal-2020-role}, we apply self-training to improve AV-NSL.
We extend Benepar to the speech domain and introduce s-Benepar, which takes segment-level continuous mean-pooling HuBERT representations, instead of words, as the input, and outputs the constituency parse trees.

\subsection{Unsupervised Decoding}
\label{subsec: unsup_decoding}
Another key ingredient of AV-NSL is applying consistency-based decoding \cite{shi2019visually}, which is similar in spirit to minimum Bayes risk (MBR) decoding, for both spoken word segmentation and phrase-structure induction.
Given a loss function $\ell_\textit{MBR}(O_1, O_2)$ between two outputs $O_1$ and $O_2$, and a set of $k$ outputs $\mathcal{O} = \{O_1, \ldots, O_k\}$, we select the optimal output 
$$
\hat{O} = \arg\min_{O'\in \mathcal{O}} \sum_{O''\in \mathcal{O}} \ell_\textit{MBR}(O', O'').
$$
For word segmentation, we define the loss between two segmentation proposals $\mathcal{S}_1$ and $\mathcal{S}_2$ as 
$\ell_\textit{MBR}(\mathcal{S}_1, \mathcal{S}_2) = -\textsc{mIoU}(\mathcal{S}_1, \mathcal{S}_2),$ where $\textsc{mIoU}(\cdot,\cdot)$ denotes the mean intersection over union ratio across all matched pairs of predicted word spans.
We match the predicted word spans using the maximum weight matching algorithm 
 \cite{10.1145/6462.6502}, where word spans correspond to vertices, and we define edge weights by the temporal overlap between the corresponding spans.

For phrase structure induction, the loss function between two parse trees $\mathcal{T}_1$ and $\mathcal{T}_2$ is  
$\ell_\textit{MBR}(\mathcal{T}_1, \mathcal{T}_2) = 1 - F_1(\mathcal{T}_1, \mathcal{T}_2),$ where $F_1(\cdot, \cdot)$ denotes the $F_1$ score between the two trees.

%% file: src/40-experiments.tex
\section{Experiments}
\label{sec: experiments}

\subsection{Setup}
\label{subsec: exp_setting}

\vspace{1mm} \noindent \textbf{Datasets.}
We first evaluate models on SpokenCOCO \cite{hsu2020text}, the spoken version of MSCOCO \cite{lin2014microsoft} where the text captions in English are read verbally by humans. 
It contains 83k/5k/5k images for training, validation and testing, respectively. 
Each image has five corresponding captions. 

We also extend our experiments to German, where we synthesize German speech from the Multi30K captions \cite{W16-3210}.\footnote{Synthesized with pre-trained German Tacotron2 from \url{https://github.com/thorstenMueller/Thorsten-Voice}.}
It contains 29k/1k/1k images for training, validation and testing, respectively. 
Each image has one corresponding caption. 
Following \cite{shi2019visually}, we use pretrained Benepar \cite{kitaev-klein-2018-constituency}, an off-the-shelf parser, to generate the oracle parse trees for captions.
\vspace{2pt}

\vspace{1mm} \noindent \textbf{Preprocessing.}
For oracle word segmentation, we use the Montreal Forced Aligner \cite{mcauliffe2017montreal} trained on the specific language (i.e., English or German). 
We remove utterances that have mismatches between ASR transcripts and text captions. 

\vspace{-2mm}
\subsection{Baselines and Toplines}
\label{subsec: baselines}
We consider the following baselines and modeling alternatives to examine each component of AV-NSL:
\vspace{2pt}

\vspace{1mm} \noindent \textbf{Trivial tree structures.} Following \cite{shi2019visually}, we include baselines without linguistic information: random binary trees, left-branching binary trees, and right-branching binary trees.
\vspace{2pt}

\vspace{1mm} \noindent \textbf{AV-cPCFG.} We train compound probabilistic context free grammars (cPCFG) \cite{kim2019compound} on word-level discrete speech tokens given by VG-HuBERT.
Unlike in AV-NSL, the segment representations are discretized via k-Means to obtain word-level indices; that is, AV-cPCFG leverages visual cues only for segmentation and segment representations, and not for phrase structure induction. 
\vspace{2pt}

\vspace{1mm} \noindent \textbf{DPDP-cPCFG.} In contrast to AV-cPCFG, DPDP-cPCFG does not rely on any visual grounding throughout. 
We use DPDP \cite{kamper2022word} and pre-trained HuBERT \cite{hsu2021hubert} followed by k-Means to obtain discrete word indices.\footnote{We sweep the number of word clusters over $\{1\text{k}, 2\text{k}, 4\text{k}, 8\text{k}, 12\text{k}, 16\text{k}\}.$} 
\vspace{2pt}

\vspace{1mm} \noindent \textbf{Oracle AV-NSL} (topline). To remove the uncertainty of unsupervised word segmentation, we directly train AV-NSL on top of oracle word segmentation via forced alignment. 
Due to the absence of VG-HuBERT, the frame-level representations $R$ are obtained from pre-trained HuBERT while the attention weights $a_{i,t}$ are parameterized by a 1-layer MLP, jointly trained with the tree sampling module instead. 

\begin{table*}[!ht]
\centering
\begin{center}
\scalebox{0.925}{
    \begin{tabular}{lllcc}
    \toprule
    \multicolumn{3}{c}{\bf Model} & {\bf Output} & \multirow{2}{*}{\bf \textsc{SAIoU}} \\ 
    \cmidrule(lr){1-3}
    {\bf Syntax Induction} & {\bf Segmentation} & {\bf Seg. Representation (continuous/discrete)} & {\bf Selection} & {} \\
    \midrule
    Right-Branching & VG-HuBERT+MBR$_{10}$ &  & & \textbf{0.546} \\ 
    Right-Branching & DPDP & & & 0.478 \\
    \cdashlinelr{1-5}
    AV-cPCFG & VG-HuBERT+MBR$_{10}$ & VG-HuBERT$_{10}$+4k km (discrete) & last ckpt. (supervised) & 0.499 \\
    AV-cPCFG & VG-HuBERT+MBR$_{10}$ & VG-HuBERT$_{10}$+8k km (discrete) & last ckpt. (supervised) & 0.481 \\
    \cdashlinelr{1-5}
    DPDP-cPCFG & DPDP & HuBERT$_{2}$+2k km (discrete) & last ckpt. (supervised) & 0.465 \\ 
    \cdashlinelr{1-5}
    AV-NSL & VG-HuBERT+MBR$_{10}$ & VG-HuBERT$_{10}$ (continuous) & MBR over 10\textsuperscript{th} layer & 0.516 \\ 
    AV-NSL & VG-HuBERT+MBR$_{10}$ & VG-HuBERT$_{10,11,12}$ (continuous) & MBR over $\{10\textsuperscript{th}, 11\textsuperscript{th}, 12\textsuperscript{th}\}$ layer & 0.521 \\ 
    \bottomrule
    \end{tabular}
}
\vspace{-10pt}
\caption{Fully-unsupervised English phrase structure induction results on SpokenCOCO.
Subscripts denote layer number, e.g. HuBERT$_{10}$ denotes the 10\textsuperscript{th} layer representation from HuBERT.
We list the best-performing hyperparameters for each modeling choice. 
}
\label{tab: main_result}
\end{center}
\end{table*}

\vspace{-2mm}
\subsection{Evaluation Metrics}
\label{subsec: eval}

\vspace{1mm}\noindent \textbf{Word segmentation.} We use the standard word boundary prediction metrics (precision, recall and $F_1$), which are calculated by comparing the temporal position between inferred word boundaries and forced aligned word boundaries. 
An inferred boundary located within $\pm 20$\textit{ms} of a forced aligned boundary is considered a successful prediction.
\vspace{2pt}

\vspace{1mm}\noindent \textbf{Parsing.} For parsing with oracle word segmentation, we use \textsc{ParsEval} \cite{black-etal-1991-procedure} to calculate the $F_1$ score between the predicted and reference parse trees. 
For parsing with inferred word segmentation, due to the mismatch in the number of nodes between the predicted and reference parse trees, we use the structured average intersection-over-union ratio (\textsc{SAIoU} \cite{shi-etal-2023-structure}) as an additional metric.

\textsc{SAIoU} takes both word segmentation quality and temporal overlap between induced constituents into consideration. 
Concretely, the input is two constituency parse trees over the same speech utterance, $\mathcal{T}_1=\{a_i\}_{i=1}^n$ and $\mathcal{T}_2=\{b_j\}_{j=1}^m$, where $a_i$ and $b_j$ are time spans. 
Suppose $a_i$ from $\mathcal{T}_1$ is aligned to $b_j$ from $\mathcal{T}_2$.  In a valid alignment, the following conditions must be satisfied: (1) any descendant of $a_i$ may either align to a descendant of $b_j$ or be left unaligned; (2) any ancestor of $a_i$ may either align to an ancestor of $b_j$ or be left unaligned; (3) any descendant of $b_j$, may either align to a descendant of $a_i$ or be left unaligned; (4) any ancestor of $b_j$, may either align to an ancestor of $a_i$ or be left unaligned.

Given a Boolean matrix $\bm{A}$, where $A_{i,j}=1$ denotes that $a_i$ aligns to $b_j$, we compute the structured average \textsc{IoU} between $\mathcal{T}_1$ and $\mathcal{T}_2$ over $\bm{A}$ by 
$$\textsc{SAIoU}(\mathcal{T}_1, \mathcal{T}_2; \bm{A}) = \frac{2}{n+m} \left(\sum_{i=1}^{n_1} \sum_{j=1}^{n_2} A_{i,j} \textsc{IoU}(a_i, b_j)\right),$$
and the final evaluation result is obtained by maximizing the \textsc{SAIoU} score across all valid alignments. 
The calculation of the optimal \textsc{SAIoU} score can be done within $\mathcal{O}(n^2m^2)$ time by dynamic programming. 

\begin{table}[!t]
\begin{center}
\scalebox{0.95}{
    \begin{tabular}{cccccc}
    \toprule
    \bf Method & \bf Decoding   & \bf Precision & \bf Recall & \bf $F_1$ \\
    \midrule
    DPDP \cite{kamper2022word} &supervised & 17.37&9.00&11.85 \\
    \cdashlinelr{1-6}
    VG-HuBERT \cite{peng2022word} & supervised       &\bf 36.19&27.22&31.07 \\
    \cdashlinelr{1-6}
    VG-HuBERT &  supervised &34.34 &29.85 &31.94 \\
    w/ seg. ins. (ours) & MBR     & 33.31 &\bf 34.90 &\bf 34.09 \\
    \bottomrule
    \end{tabular}
}
\vspace{-3mm}
\caption{English word segmentation results on the SpokenCOCO validation set. 
Supervised decoding methods require an annotated development set to choose the best hyperparameters.
The best number in each column is in boldface. 
VG-HuBERT with segment insertion and MBR decoding achieves the best boundary $F_1$.}
\label{tab:word_seg_f1}
\end{center}
\end{table}

\subsection{Unsupervised Word Segmentation}
\label{subsec: unsup_word_seg}
We validate the effectiveness of our unsupervised word segmentation approach. 
We first compare our improved VG-HuBERT with segment insertion to the original VG-HuBERT \cite{peng2022word} and DPDP \cite{kamper2022word}, a speech-only word segmentation method (Table~\ref{tab:word_seg_f1}). 
We find that segment insertion improves recall and hurts precision, and achieves the highest $F_1$ score.

Next, we compare MBR-based and supervised decoding. 
For efficiency in practice, we implement MBR-based decoding as follows: we first run a pilot hyperparameter selection, performing word segmentation on all candidates in the SpokenCOCO validation set, and subsequently choose the $10$ most selected sets of hyperparameters to perform another round of MBR selection on the training set.

For German word segmentation, we employ identical models and settings as those used for English, as \cite{peng2023syllable} has shown that the word segmentation capability of English VG-HuBERT demonstrates cross-lingual generalization without any adaptation. On German Multi30K, our method achieves an $F_1$ score of $37.46$ with MBR, which outperforms that of supervised hyperparameter tuning ($36.45$).

\subsection{Unsupervised Phrase Structure Induction}
\label{subsec: unsup_syntax}
We quantitatively show that AV-NSL learns meaningful phrase structure given word segments (Table~\ref{tab: main_result}). 
The best performing AV-NSL is based on our improved VG-HuBERT with MBR top 10 selection for word segmentation, VG-HuBERT layers as the segment representations, and another MBR decoding over phrase structure induction hyperparameters, including training checkpoints and segment representation layers.
Comparing AV-NSL against AV-cPCFG and AV-cPCFG against DPDP-cPCFG, we empirically show the necessity of training AV-NSL on \textit{continuous} segment representation instead of discretized speech tokens, and the effectiveness of visual grounding in our overall model design. 

\begin{table}[t]
\centering
\begin{center}
    \begin{tabular}{lcc}
    \toprule
    \bf Segment Representation & \bf Output Selection & \bf \textsc{SAIoU} \\
    \midrule
    HuBERT & last ckpt. & \textbf{0.538} \\
    {HuBERT$_{2,4,6,8,10,12}$} & MBR & 0.536 \\
    \bottomrule
    \end{tabular}
\vspace{-10pt}
\caption{Results of self-training with s-Benepar, trained on outputs from the best AV-NSL model (\textsc{SAIoU} 0.521) from Table~\ref{tab: main_result}.
Inputs to s-Benepar are segment-level HuBERT representations instead of VG-HuBERT representations.
}
\label{tab: self_train_result}
\end{center}
\end{table}

Next, we compare the performance of AV-NSL with and without self-training (Table~\ref{tab: self_train_result}), and find that self-training with an s-Benepar backbone improves the best AV-NSL performance from 0.521 (Table~\ref{tab: main_result}) to 0.538. 

Thirdly, Table~\ref{tab: oracle_seg_result} isolates phrase structure induction from word segmentation quality with oracle AV-NSL. 
Unlike in Table~\ref{tab: main_result}, we can adopt \textsc{ParsEval} $F_1$ score \cite{black-etal-1991-procedure} for evaluation since there is no mismatch in the number of tree nodes.
With proper segment-level representations, unsupervised oracle AV-NSL matches or out-performs text-based VG-NSL.
Similarly to Table~\ref{tab: self_train_result}, self-training with s-Benepar on oracle AV-NSL trees further improves the syntax induction results, almost matching that of right-branching trees. 

\begin{table}[!]
\begin{center}
\scalebox{0.8}{
    \begin{tabular}{llcc}
    \toprule
    \multicolumn{2}{c}{\bf Model} & {\bf Output} & \multirow{2}{*}{$F_1$} \\ 
    \cmidrule(lr){1-2}
    {\bf Syntax Induction} & {\bf Seg. Representation} & {\bf Selection} & {} \\
    \midrule
    Right-Branching & N/A & N/A & \textbf{57.39} \\
    \cdashlinelr{1-4}
    VG-NSL & word embeddings & Supervised & 53.11 \\
    oracle AV-NSL & HuBERT$_{2}$ & Supervised & 55.51 \\ 
    \cdashlinelr{1-4}
    oracle AV-NSL $\rightarrow$ s-Benepar & HuBERT$_{2}$ & MBR & 57.24 \\
    \bottomrule
    \end{tabular}
}
\vspace{-2mm}
\caption{\textsc{ParsEval} $F_1$ scores given oracle segmentation. The best number is in boldface. 
}
\label{tab: oracle_seg_result}
\end{center}
\end{table}

Perhaps surprisingly, right-branching trees (RBT) with oracle and VG-HuBERT word segmentation reach the best English \textsc{SAIoU} and $F_1$ scores on SpokenCOCO, respectively. 
We note that the RBTs highly align with the head-initiality of English \cite{baker2001atoms}, especially in our setting where all punctuation marks were removed.
In contrast, our experiments on German show that AV-NSL out-performs both RBTs and left-branching trees in terms of \textsc{SAIoU} (Table~\ref{tab: german_result}).\footnote{For German grammar induction with oracle segmentation, oracle AV-NSL attains 33.94 $F_1$ while LBT/RBT attain 26.70/25.30 $F_1$ respectively.}

\begin{table}[!t]
\centering
\begin{center}
\scalebox{0.9}{
    \begin{tabular}{llcc}
    \toprule
    \multicolumn{2}{c}{\bf Model} & {\bf Output} & \multirow{2}{*}{\bf \textsc{SAIoU}} \\ 
    \cmidrule(lr){1-2}
    {\bf Induction} & {\bf Segmentation} & {\bf Selection} & {} \\
    \midrule
    Right-Branching & VG-HuBERT+MBR$_{10}$ & N/A & 0.456 \\ 
    Left-Branching & VG-HuBERT+MBR$_{10}$ & N/A & 0.461 \\
    \cdashlinelr{1-4}
    AV-NSL & VG-HuBERT+MBR$_{10}$ &  MBR & \textbf{0.487} \\ 
    \bottomrule
    \end{tabular}
}
\vspace{-10pt}
\caption{Phrase structure induction results on the German Multi30K test set. The best number is in boldface. 
}
\label{tab: german_result}
\end{center}
\end{table}

%% file: src/50-analysis.tex
\subsection{Analyses}
\label{subsec: analysis}

\vspace{1mm}\noindent \textbf{Unsupervised Constituent Recall:} Following \cite{shi2019visually}, we show the recall of specific types of constituents (Table~\ref{tab: recall_analysis}).
While VG-NSL benefits from the head-initial (HI) bias, where abstract words are encouraged to appear in the beginning of a constituent, AV-NSL outperforms all variations of VG-NSL on all constituent categories except NP. 
\vspace{2pt}

\vspace{1mm} \noindent \textbf{Ablation Study:}
We introduce three ablations to evaluate the efficacy of high-quality word segmentation, visual representation, and speech representation (Table~\ref{tab: quality_result}).
Concretely, we train AV-NSL with the following modifications: 
\vspace{-0mm}
\begin{enumerate}[leftmargin=*,topsep=4pt]
    \setlength{\itemsep}{-1pt}
    \item Given the number of words $n$, we divide the speech utterances uniformly into $n$ chunks to get the word segmentation, and use the same visual representations as AV-NSL.
    \item We replace visual representations with random vectors, where each pixel is independently sampled from a uniform distribution, and use the oracle word segmentation.
    \item We replace the self-supervised speech representations (HuBERT) with log-Mel spectrograms. 
\end{enumerate}
\vspace{-0mm}
We observe significant performance drops in all settings, compared to oracle AV-NSL. 
This set of results complements Table~\ref{tab: main_result}, stressing that precise word segmentation and both high-quality visual and speech representations are all necessary for phrase structure induction from speech.

\begin{table}[]
\vspace{-0.01in}
\small
  \begin{center}
    \scalebox{1}{
    \begin{tabular}{lccccc}
            \toprule
            \multirow{2}{*}{\bf Model} & \multirow{2}{*}{$F_1$} & \multicolumn{4}{c}{\bf Constituent Recall} \\ \cmidrule(lr){3-6}
            & & \bf NP & \bf VP & \bf PP & \bf ADJP \\
            \midrule
            VG-NSL~\cite{shi2019visually} & 50.4 & \textbf{79.6} & 26.2 & 42.0 & 22.0 \\ 
            VG-NSL + HI & 53.3 & 74.6 & 32.5 & 66.5 & 21.7 \\ 
            VG-NSL + HI + FastText & 54.4 & 78.8 & 24.4 & 65.6 & 22.0 \\ 
            \cdashlinelr{1-6}
            oracle AV-NSL & \textbf{55.5} & 55.5 & \textbf{68.1} & \textbf{66.6} & \textbf{22.1} \\
            \bottomrule
    \end{tabular}
    }
    \vspace{-10pt}
  \end{center}
\caption{Recall of specific typed phrases, incl. noun phrases (NP), verb phrases (VP), prepositional phrases (PP) and adjective phrases (ADJP), and overall $F_1$ score, evaluated on SpokenCOCO test set. 
VG-NSL numbers are taken from \cite{shi2019visually}.}
\label{tab: recall_analysis}

\end{table}
  
\begin{table}[]
  \small
  \begin{center}
  \scalebox{1}{%
  \begin{tabular}{llcc}
            \toprule
            \multicolumn{2}{c}{\bf Model} & \multirow{2}{*}{\bf Visual} & \multirow{2}{*}{$F_1$} \\ 
            \cmidrule(lr){1-2}
            {\bf Word Segmentation} & {\bf Seg. Repre.} & {} & \\
            \midrule
            MFA & HuBERT$_2$ & ResNet 101 & 55.51\\
            \cdashlinelr{1-4}
            Uniform & HuBERT$_{2}$ & ResNet 101 & 48.97 \\ 
            MFA & HuBERT$_2$ & random & 31.23 \\
            MFA & logMel spec & ResNet 101 & 42.01 \\
            \bottomrule
            \end{tabular}}
  \end{center}
  
      \caption{\textsc{ParsEval} $F_1$ scores for ablations over word segmentation, visual representation, and speech representation.}
      \label{tab: quality_result}
  \vspace{-.1in}
  \end{table}

%% file: src/60-conclusion.tex
\section{Conclusion and Discussion}
Previous research has achieved notable progress in zero-resource speech processing and grammar induction by employing multi-modal techniques. 
In our study, we propose an approach to model human language acquisition that leverages the visual modality to acquire language competence. 
Our approach, AV-NSL, encompasses the extraction of word-level representations from speech and the derivation of syntactic structures from those representations, thereby eliminating the reliance on text. 
Through quantitative and qualitative analyses, we demonstrate on both English and German that our proposed model successfully infers meaningful constituency parse trees based on continuous word segment representations.
Our work represents the initial step in grammar induction within textless settings, paving the way for future research endeavors, which include but are not limited to (1) building end-to-end models that take spoken utterances and produce their syntactic analysis, (2) understanding the contribution of various grounding signals to grammar induction, and (3) modeling human language acquisition in grounded environments.  